\documentclass[letterpaper, 10 pt, conference]{ieeeconf}  

\IEEEoverridecommandlockouts                              

\usepackage{graphicx} 
\usepackage{amsmath} 
\usepackage[normalem]{ulem}
\usepackage{multirow, booktabs, makecell, float}
\usepackage{gensymb}
\usepackage[table,xcdraw]{xcolor}

\usepackage{pifont,xcolor}
\usepackage{cite}

\title{\LARGE \bf
	3DCarGen: Scalable 3D Car Generation via 3D-consistent Multi-view Synthesis
}

\author{
	Hongli Xiao$^{1,2*}$, Youjian Zhang$^{3*}$,  Yaohui Jin$^{1}$,   Xiaoguang Ren$^{2}$,  Wenjing Yang$^{4}$, Long Lan$^{4\dagger}$
	\thanks{*Equal contribution. }
	\thanks{$\dagger$Corresponding author.}
	\thanks{$^{1}$Shanghai Jiao Tong University, China (\{honglixiao, jinyh\}@sjtu.edu.cn)}%
	\thanks{$^{2}$Academy of Military Science, China (rxg\_nudt@126.com)}%
	\thanks{$^{3}$The University of Sydney, Australia (zyj8813@gmail.com)}%
	\thanks{$^{4}$College of Computer Science and Technology, National University of Defense Technology, China (\{wenjing.yang, long.lan\}@nudt.edu.cn)}%
	\thanks{
		This work was supported by the National Natural Science Foundation of China (No. 62376282) and the Science and Technology Innovation Program of Hunan Province (No.2025RC3117).}
}

\begin{document}

	\maketitle
	\thispagestyle{empty}
	\pagestyle{empty}

	\begin{abstract}
		High-quality 3D vehicle assets are essential for autonomous driving simulation. Although multi-view diffusion-based paradigms enable controllable single-image reconstruction, they typically produce limited viewpoints and exhibit cross-view geometric inconsistencies, thereby reducing reconstruction fidelity in real-world scenarios.
		In this work, we introduce \textbf{3DCarGen}, a scalable single-view 3D car generation framework designed for real-world images by synthesizing an arbitrary number of 3D-consistent multi-view images.
		Specifically, given a single image as input, we first synthesize a set of images from fixed viewpoints. These images are then fed into a feed-forward reconstruction model, resulting in a coarse 3D representation based on 3D Gaussian Splatting.
		Conditioned on this explicit 3D prior, our multi-view diffusion model generates 3D-consistent images from arbitrary camera viewpoints.
		We further extend a fast mesh reconstruction algorithm by incorporating color-normal joint optimization to recover detailed and coherent 3D vehicle models from the synthesized dense views.
		Extensive experiments on synthetic and real-world datasets demonstrate that our approach achieves robust geometric consistency and reconstruction fidelity compared to existing methods.
		Project page: https://honglixiao.github.io/3dcargen.github.io/.

	\end{abstract}

	\section{INTRODUCTION}
	\label{sec:intro}

	
	3D car models play a critical role in many real-world applications, including autonomous driving, scene understanding, digital twins, and synthetic data generation.
	Traditionally, they are manually crafted by professionals using specialized tools, which is time-consuming and labor-intensive, particularly when modeling vehicles from real-world images taken in unrestricted environments. Consequently, there is a growing need for efficient methods to automatically generate high-quality 3D car models.
	
	In recent years, several approaches have been proposed for automatic 3D car generation~\cite{liu2024car, du2024dreamcar, lin2024drive, chen2024rgm}. 
	Though effective to some extent, these methods either demand high computational costs~\cite{liu2024car} or require multiple input views~\cite{du2024dreamcar, wang2023cadsim}, yet the quality of the generated 3D car assets still lacks realism and fine details. Thus, their scalability and applicability in real-world scenarios are still limited.
	
	Recent advances in diffusion-based image-to-3D generation~\cite{shi2023zero123++, xu2024instantmesh, long2024wonder3d, wu2024unique3d} have enabled automatic reconstruction from a single image using a two-stage paradigm including multi-view synthesis followed by 3D reconstruction. 
	Compared with previous optimization-based methods~\cite{pooledreamfusion, lin2023magic3d, chen2023fantasia3d, wang2024prolificdreamer}, the feed-forward multi-view (MV) diffusion pipeline offers higher efficiency and controllability, making it promising for practical applications.

	\begin{figure}[t]
		\begin{center}
			\includegraphics[width=0.85\linewidth]{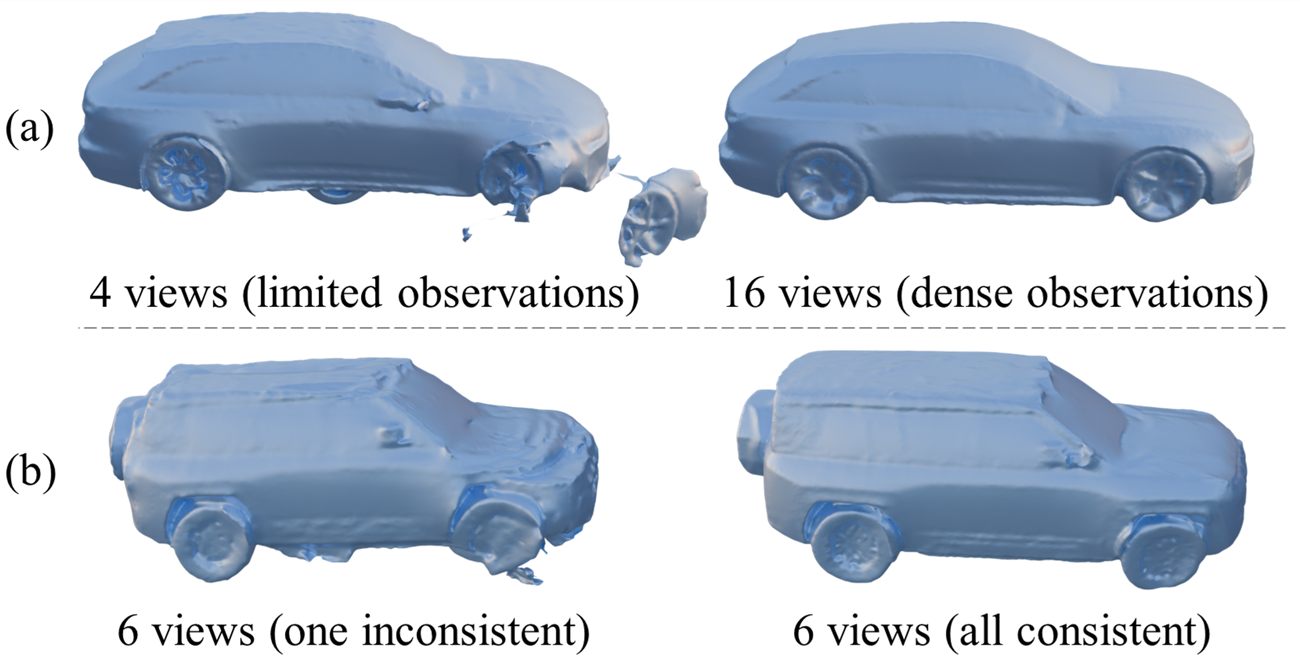}
		\end{center}
		\vspace{-4mm}
		\caption{
			Effect of view coverage and cross-view consistency on mesh reconstruction using the ISOMER~\cite{wu2024unique3d} algorithm.
			(a) Reconstruction results using different numbers of input views. 
			Sparse views (left) lead to degraded reconstruction accuracy, while dense views (right) produce more reliable results.
			(b) Reconstruction results with the same number of views but different degrees of cross-view consistency.
			Even a single inconsistent view can severely degrade the reconstructed geometry.
		}
		\label{fig:2issues_of_MVDiffusion}
		\vspace{-6mm}
	\end{figure}
	
	However, existing MV diffusion models typically generate a fixed number of images from predefined viewpoints (usually 4-6 images for generation efficiency) and often exhibit varying degrees of local inconsistency across synthesized views. 
	The limited number of generated views often leads to inaccurate or blurred reconstructions in regions with insufficient observations (Fig.~\ref{fig:2issues_of_MVDiffusion} (a)). 
	Furthermore, cross-view inconsistency can significantly degrade the reconstruction quality, as most reconstruction algorithms assume accurate geometric alignment between multi-view observations (Fig.~\ref{fig:2issues_of_MVDiffusion} (b)).
	The above issues become more pronounced when applied  to real-world images. 
	Many existing methods~\cite{long2024wonder3d, shi2023MVDream} assume orthographic input views or  operate in an orthographic camera space, whereas real-world images are typically captured by perspective cameras. 
	This mismatch often results in distorted or geometrically inconsistent novel views during multi-view generation. 
	In addition, some approaches~\cite{wu2024unique3d, long2024wonder3d} prefer front-facing input images and perform poorly when the object is viewed from other angles, while vehicles in real-world scenes may appear from arbitrary viewing directions.

	To address the aforementioned problems, we propose 3DCarGen, a scalable framework for generating high-quality 3D car models from a single real-world image. 
	The key idea is to synthesize dense multi-view images from arbitrary viewpoints to provide sufficient visual coverage for reliable mesh reconstruction. 
	To ensure geometric consistency across views, we introduce a coarse 3D representation as a synchronization condition during multi-view synthesis.
	Our method mainly consists of two parts, 3D-consistent multi-view synthesis and fast mesh reconstruction.

	Specifically, given a single input image, we first adopt a MV diffusion model~\cite{shi2023zero123++} that supports perspective image input to generate a set of generally consistent views with fixed camera parameters and predefined angles.
	Inspired by~\cite{chen20243d, wu2024reconfusion}, we train a feed-forward reconstruction model to serve as an \textit{input-sync} for subsequent 3D-consistent multi-view synthesis.
	It takes the fixed views as input, aggregates features via cross-view attention, and predicts a unified 3D Gaussian splatting (3DGS).
	The rendering of the 3DGS model can provide coarse, blurred yet geometrically accurate images.
	Then we train a multi-view diffusion model to achieve arbitrary viewpoint generation conditioned on the coarse 3D representation, ensuring geometric consistency across views.
	During inference, we further incorporate a noising-denoising process into the model to enhance coherence, inspired by SDEdit~\cite{mengsdedit}.


	
	Finally, with sufficient multi-view color images and normal maps, we can obtain high-fidelity vehicle meshes through a fast optimization-based reconstruction stage.
	Building upon the ISOMER~\cite{wu2024unique3d} framework, we extend it to support an arbitrary number of specified views, namely ISOMER+. 
	Different from ISOMER which optimizes the geometry with only normal maps, we perform a joint optimization with both color images and normal maps, yielding a more accurate and smoother surface reconstruction.
	With dense views available, we further adopt a one-shot color projection strategy for vertex colorization, reducing blurry  and ghosting artifacts in the final textured mesh.

	
	Extensive experiments on both synthetic datasets and real-world images demonstrate that 3DCarGen produces 3D-consistent multi-view images and high-quality 3D car assets.
	Our main contributions are summarized as follows:
	\begin{itemize}
		\item 
		We propose 3DCarGen, a scalable framework for single image-to-3D car generation, including 3D-consistent multi-view generation and fast mesh reconstruction. Our method can effectively and robustly generate high-quality meshes from real-world vehicle images.
		
		\item 
		We propose a 3D-consistent multi-view diffusion model by introducing a coarse 3D representation as synchronization, thereby overcoming the limitations of view inconsistency and fixed viewpoints in existing methods.
		
		\item 
		We extend ISOMER~\cite{wu2024unique3d} to ISOMER+, enabling fast and robust mesh reconstruction from an arbitrary number of multi-view images, producing meshes with high-quality geometry and textures.
		
		
		
	\end{itemize}

	\section{RELATED WORKS}
	\label{sec:related_works}
	
	\subsection{General 3D Generation}
	In recent years,  image-to-3D  generation has drawn significant attention as advances in diffusion models~\cite{rombach2022high} and large-scale 3D datasets~\cite{deitke2023objaverse, deitke2023objaverse_xl}. 
	Early approaches~\cite{pooledreamfusion, lin2023magic3d, qianmagic123, chen2023fantasia3d} typically rely on per-shape optimization by distilling priors from 2D generative models. 
	Zero123~\cite{liu2023zero} has revealed the potential of fine-tuning the pre-trained Stable Diffusion model for novel view synthesis but suffers from severe view-inconsistency issues since it generates each view independently.
	Subsequent works proposed multi-view diffusion models~\cite{shi2023MVDream, wang2023imagedream} that can generate multiple images simultaneously. 
	Combined with sparse-view reconstruction methods~\cite{worchel2022multi, palfinger2022continuous, li2024m, tang2024lgm}, they further improve image-to-3D generation quality. 
	Wonder3D~\cite{long2024wonder3d} extends diffusion frameworks by jointly modeling the distributions of normal maps and color images using a multi-view cross-domain attention mechanism.
	Unique3D~\cite{wu2024unique3d} designs an instant consistent mesh reconstruction algorithm. 
	However, these two methods perform poorly for skewed or non-orthogonal inputs.
	Zero123++~\cite{shi2023zero123++} fine-tunes the Stable Diffusion model by tiling multi-views surrounding the object into a single image to handle input images with different camera intrinsic.
	Nevertheless, these models usually perform well on global semantic consistency and struggle to achieve local geometry consistency. 
	Furthermore, they can only generate a fixed number of images with predefined viewpoints, posing challenges to subsequent 3D reconstruction.
	Some \textit{input-sync} methods~\cite{chan2023generative, wu2024reconfusion} have explored incorporating 3D geometry priors by conditioning on rendered features to enhance local consistency. 
	3D-Adapter~\cite{chen20243d} adopts a 3D feedback augmentation for each denoising step via decoding and re-encoding but introduces substantial computation overhead. 
	In this work, we address these issues by incorporating a coarse 3D representation for 3D-consistent multi-view synthesis and generate sufficient number of images/normals.


	\subsection{ Car-specific 3D Generation }
	While general 3D generation works demonstrate impressive capabilities in object-level synthesis, their performance for generating real-world cars remains unsatisfactory.
	GINA-3D~\cite{shen2023gina} introduces a transformer-based generative model that uses driving data from both the camera and LiDAR to create 3D implicit neural assets. 
	DreamCar~\cite{du2024dreamcar} proposes a four-stage 3D car reconstruction method leveraging car-specific prior given one to five supervision images. 
	Its performance significantly degrades when only single-view inputs are available.
	Car-Studio~\cite{liu2024car} utilizes an encoder-decoder based network to learn priors and regresses NeRF representations from single car images. 
	However, these works suffer from high training and rendering costs and often exhibit artifacts upon large view changes due to overfitting.
	Drive-1-to-3~\cite{lin2024drive} constructs a vehicle-specific multi-view diffusion model fine-tuned from Free3D~\cite{zheng2024free3d}. 
	RGM~\cite{chen2024rgm} proposes a relightable 3DGS generative model for generating 3D car assets with material properties. 
	Nevertheless, its reliance on a self-curated dataset may limit generalization to real-world cars.

	\begin{figure*}[htbp]
		\begin{center}
			\includegraphics[width=\linewidth]{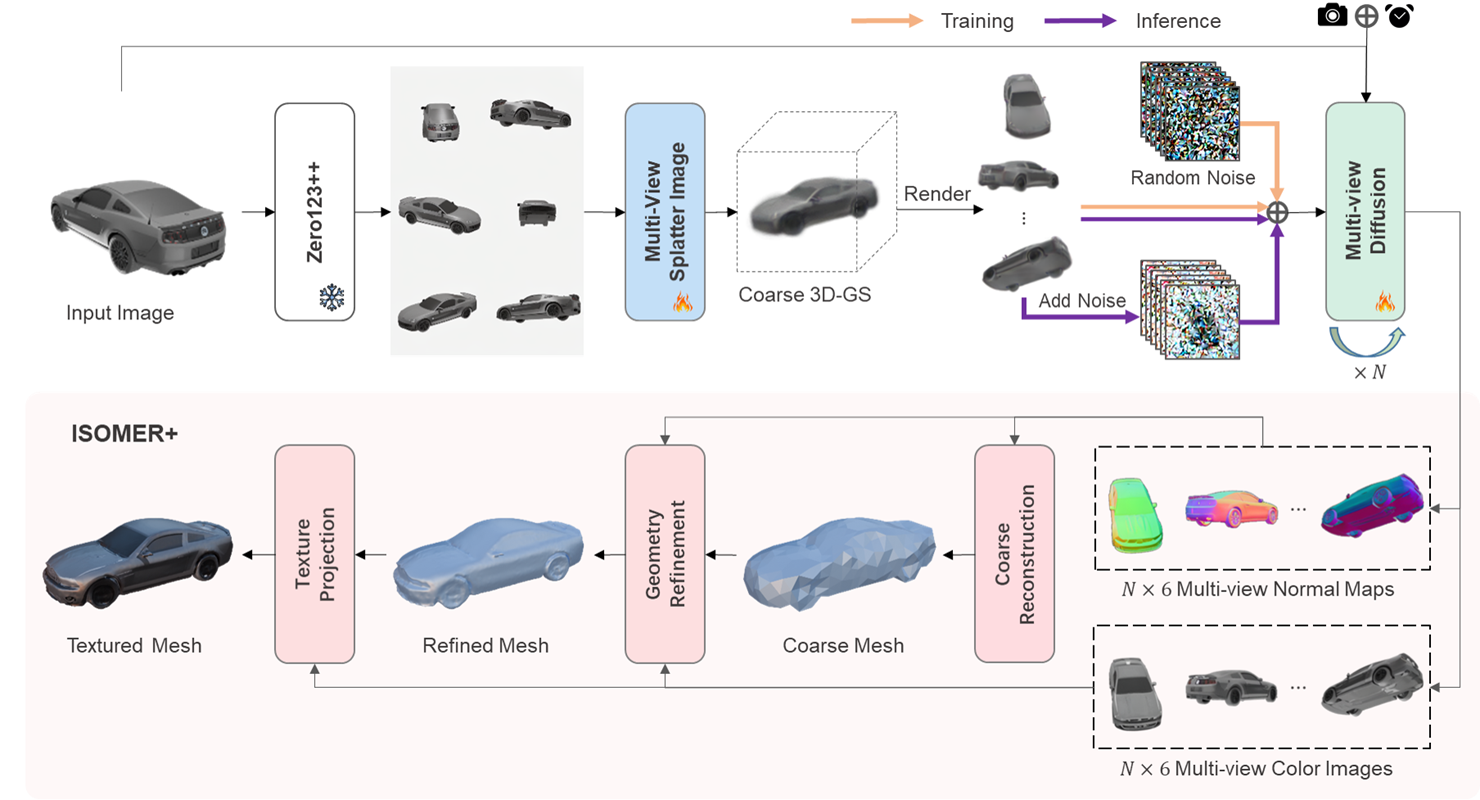}
		\end{center}
		\vspace{-4mm}
		\caption{
			\textbf{Overview of our proposed 3DCarGen pipeline. }
			The pipeline mainly consists of two stages: (1) Top: 3D-consistent multi-view generation through 3D synchronization. (2) Bottom: fast 3D reconstruction by ISOMER+.   
		}
		\label{fig:pipeline}
		\vspace{-4mm}
	\end{figure*}

	\section{METHODS}
	\label{sec:methods}
	
	
	3DCarGen is a two-stage method for single-view 3D car generation. The overall pipeline is illustrated in Fig.~\ref{fig:pipeline}. 
	In the first stage, we utilize a synchronized 3D representation to generate sufficient and 3D-consistent color images and normal maps from different viewpoints.
	Second, we extend a fast remeshing algorithm~\cite{wu2024unique3d} to recover the textured mesh, called ISOMER+. 

	\subsection{Preliminary: Feed-forward 3D Gaussians}
	\label{subsec:method_preliminary}

	Building upon 3DGS~\cite{liu20243dgs}, Splatter Image (SI)~\cite{szymanowicz2024splatter} introduces an image-to-image network architecture $S$ for monocular and sparse-view object-level 3D reconstruction. It employs the auto-encoder structure to learn 3D priors and predicts per-pixel Gaussian parameters $\{ \mathbf{g}_i \}_{i=1}^{H \times W }$ for the input image $I$ in a feed-forward manner. 
	For multiple input images $\{ I_j \}_{j=1}^{N }$, SI can obtain a composite mixture of 3D Gaussians simply by taking their union \( \Theta = \bigcup_{j=1}^N \phi_j[S(I_j)] \) according to their corresponding angle difference  \( \phi \) from the first input.
	


	\subsection{3D-consistent Multi-view Generation through 3D Synchronization}
	\label{subsec:method_mv_gene}
	
	Take the formulation of Wonder3D~\cite{long2024wonder3d} as an example, the current multi-view diffusion usually takes an input image $I^{in}$ and a collection of target poses $ \Pi ^{tgt} = \{ \pi_j ^{tgt} \}_{j=1}^{K}$ as input, and learn the joint distribution (denoted as $p(\mathbf{z}))$ over the output color images $ \mathbf{I}^{tgt} = \{ I_j^{tgt} \}_{j=1}^{K} $ and normal maps $ \mathbf{N}^{tgt} = \{ N_j^{tgt} \}_{j=1}^{K} $ at some specific target views:
	\begin{equation}
		p(\mathbf{z}) = p(\mathbf{I}^{tgt}, \mathbf{N}^{tgt} | I^{in}, \Pi ^{tgt} ).
		\label{eq:placeholder_label}
	\end{equation}
	The alignment between camera poses and output images relies solely on the condition of target poses $\Pi ^{tgt}$. 
	While works well for specific and limited target views, it becomes challenging when views are arbitrarily chosen.
	
	\noindent \textbf{Generation of coarse 3D representation.}
	Our solution is to generate a coarse 3D representation to synchronize the fine-grained generation process.
	First, to handle car images taken from the real world with varying camera intrinsic and distances, we exploit the robustness of Zero123++~\cite{shi2023zero123++} to preprocess the wild perspective images. It is robust to a wide range of input field-of-view (FOV) angles and is able to unify the output FOV angles to $30^\circ$. Given input $I^{in}$ with an arbitrary azimuth and elevation angle, Zero123++ outputs six multi-view images $\mathbf{I}^{Zero} = \{ I^{Zero}_i \}_{i=1}^{6}$ with fixed absolute elevation and relative azimuth angles $\Pi ^{Zero}$. 
	
	Then, we adopt Splatter Image (SI)~\cite{szymanowicz2024splatter} to generate a coarse 3D representation.
	However, SI with single-view input performs poorly on out-of-domain images and struggles to synthesize unseen regions.
	To address these issues, we train a multi-view Splatter Image (MV-SI) that produces robust 3D representations. Specifically, MV-SI takes $\mathbf{I}^{Zero}$ as input,  and leverages cross-attention to fuse information from multiple views. Meanwhile, it can mitigate 3D-inconsistencies introduced by Zero123++, resulting in a low-resolution but coherent 3D Gaussian representations $\mathbf{G}$. 
	Given the coarse 3D model, we can render geometrically consistent target views from arbitrarily specified perspectives: 
	\begin{equation}
		\mathbf{I}^{coarse} =\{ \mathcal{R}( \text{MV-SI}(\mathbf{I}^{Zero},  \Pi ^{Zero}) ,  \pi_j ^{tgt} ) \}_{j=1}^{K},
		\label{eq:placeholder_label}
	\end{equation}
	where $\mathcal{R}$ denotes the fast differentiable renderer of 3DGS.
	

	\noindent \textbf{3D-consistent multi-view diffusion.}
	The rendering images of MV-SI are well aligned with predefined angles and can provide a strong 3D condition for the multi-view generation. 
	Specifically, we build upon a multi-view generation backbone Wonder3D~\cite{long2024wonder3d} and extend the training paradigm from six fixed viewpoints to six arbitrary viewpoints.
	Different from Wonder3D, we take the coarse images rendered by MV-SI as the latent condition rather than the original input image. 
	In addition, since the camera pose of the input image is unknown, but the relative poses of the coarse images and the first output of Zero123++ are known, so we use its first output as the reference image to make sure the camera pose embedding has accurate physical meaning.
	
	Both coarse images and reference images are encoded using a VAE encoder and then concatenated with the noisy latent to form the final input latents.
	In this way, the multi-view diffusion model can easily learn the relations of different camera poses, and maintain consistency across different viewpoints even across multiple generation cycles, \textit{i.e.}, generating more than 6 images with different camera poses. To further capture high-level semantic information, we use the CLIP~\cite{radford2021learning} embedding of the input image and feed it into the U-Net via cross-attention. The revised version of our multi-view generation process can be derived as:
	\begin{equation}
		p(\mathbf{z})_{3D} = p(\mathbf{I}^{tgt}, \mathbf{N}^{tgt} | I^{in}, I^{Zero}_1, \mathbf{I}^{coarse}, \Pi ^{tgt} ).
		\label{eq:our_joint}
	\end{equation}

	In training stage, we randomly sample elevation and azimuth angles for a better generalization. During inference, we can perform multiple rounds of generation to obtain specific number of multi-view images/normals with predefined camera poses.
	To enhance the geometric consistency between the generation rounds, we adopt a multi-view SDEdit~\cite{mengsdedit} sampling strategy, and use the renderings of MV-SI as guidance via noising-denoising process.
	Specifically, we use $I^{coarse}_j$ as the guided image of target view $j$ and meticulously select an intermediate time $t_0 \in (0, 1)$, which control the extent of noise added to the guided image. According to $t_0$, we add corresponding noises to the guided images and start the denoising process from this intermediate time step. 
	This sampling strategy effectively exploits the blurry but geometrical-accurate nature of splatter image rendering.
	

	
	Finally, we perform super-resolution using pre-trained Real-ESRGAN~\cite{wang2021realesrgan} to further upscale the synthesized color images and normal maps by a factor of four, achieving a resolution of $1024 \times 1024$, which helps to recover high-quality geometry and textures in the following mesh reconstruction.
	

	
	
	
	\subsection{ISOMER+: Extended ISOMER with Sufficient Viewpoints }
	\label{subsec:method_mesh_recon}
	
	ISOMER proposed in Unique3D~\cite{wu2024unique3d}  achieves fast mesh reconstruction from four orthographic views. 
	Here, we extend it to support an arbitrary number (16 by default in our experiment setup) of perspective views with flexible viewpoints, as sufficient and geometry-consistent multi-view images can further enhance the reconstruction results.
	The extended ISOMER+ consists of three steps: coarse reconstruction, geometry refinement, and texture projection. 
	
	
	\noindent \textbf{Coarse reconstruction.} 
	The original ISOMER conducts shape initialization with Poisson reconstruction  using front and back views. 
	However, cars in real-world images are not necessarily facing frontally, and the front-rear asymmetrical mesh makes such initialization less reliable.
	Following~\cite{palfinger2022continuous}, we instead initialize the mesh from a sphere ($r=0.5$) and optimize it under multi-view normal supervision.
	The objective function is defined as:
	\begin{equation}
		\mathcal{L}_{{coarse}} = \mathcal{L}_{mask} +    \mathcal{L}_{normal} + \mathcal{L}_{Lap},
		\label{eq:placeholder}
	\end{equation}
	where $\mathcal{L}_{normal}$ denotes a $\ell_2$ loss between the rendered and generated normals, $\mathcal{L}_{mask}$ is a $\ell_2$ loss between the rendered masks and the masks derived from the generated normals, and $\mathcal{L}_{Lap}$ is a Laplace regularization encouraging surface smoothness.


	
	\begin{table*}[t]
		\centering
		\small
		\caption{Quantitative comparison of novel view synthesis on the SRN-Cars  and SketchFab-Cars datasets.}
		\vspace{-3mm}
		\label{tab:nvs}
		\renewcommand{\arraystretch}{0.9}
		\setlength{\tabcolsep}{12pt}
		\begin{tabular}{cccc ccc}
			\toprule
			\multirow[c]{2}{*}{Method} & \multicolumn{3}{c}{SRN-Cars} & \multicolumn{3}{c}{SketchFab-Cars} 
			\\ \cmidrule(lr){2-4} \cmidrule(lr){5-7}
			& PSNR$\uparrow$ & SSIM$\uparrow$ & LPIPS$\downarrow$ &PSNR$\uparrow$ & SSIM$\uparrow$ & LPIPS$\downarrow$ \\
			\midrule
			Splatter Image~\cite{szymanowicz2024splatter} & 16.65  & 0.800 &  0.191    &  15.98  &  0.786  &   0.193    \\
			Wonder3D~\cite{long2024wonder3d}  &  21.11 &   0.846   &  0.112    &   19.48    & 0.835   &      0.116   \\
			Unique3D~\cite{wu2024unique3d} &  21.60   &    0.852  &   0.110     &  20.07   & 0.839   & 0.113    \\
			Zero123~\cite{liu2023zero} & 16.78	&  0.792 & 0.151 &  16.34 &   0.783	& 0.147 \\
			Zero123-XL~\cite{liu2023zero} & 17.24   & 0.795  &  0.149 &  16.83     & 0.791 &  0.140 \\
			Zero123++~\cite{shi2023zero123++}    &   17.52     &    0.810     &     0.125    &  16.73      &    0.795     &      0.130  \\
			InstantMesh~\cite{xu2024instantmesh}   &  19.62 &   0.832  & \cellcolor{yellow!30} 0.105  & 18.41 & 0.817  & \cellcolor{yellow!30} 0.106      \\
			\midrule
			MV Splatter Image  &  \cellcolor{red!15} 22.90	 &     \cellcolor{red!15} 0.866	   &        0.133    &     \cellcolor{yellow!30}  21.55  &   \cellcolor{yellow!30}   0.854    &    0.137    \\
			Ours & \cellcolor{yellow!30} 22.82     &  \cellcolor{yellow!30} 0.863 & \cellcolor{red!15} 0.097      & \cellcolor{red!15} 21.58 & \cellcolor{red!15} 0.855 & \cellcolor{red!15} 0.096   \\
			\bottomrule
		\end{tabular}
	\end{table*}
	
	\begin{figure*}[t]
		\begin{center}
			\includegraphics[width=0.95\linewidth]{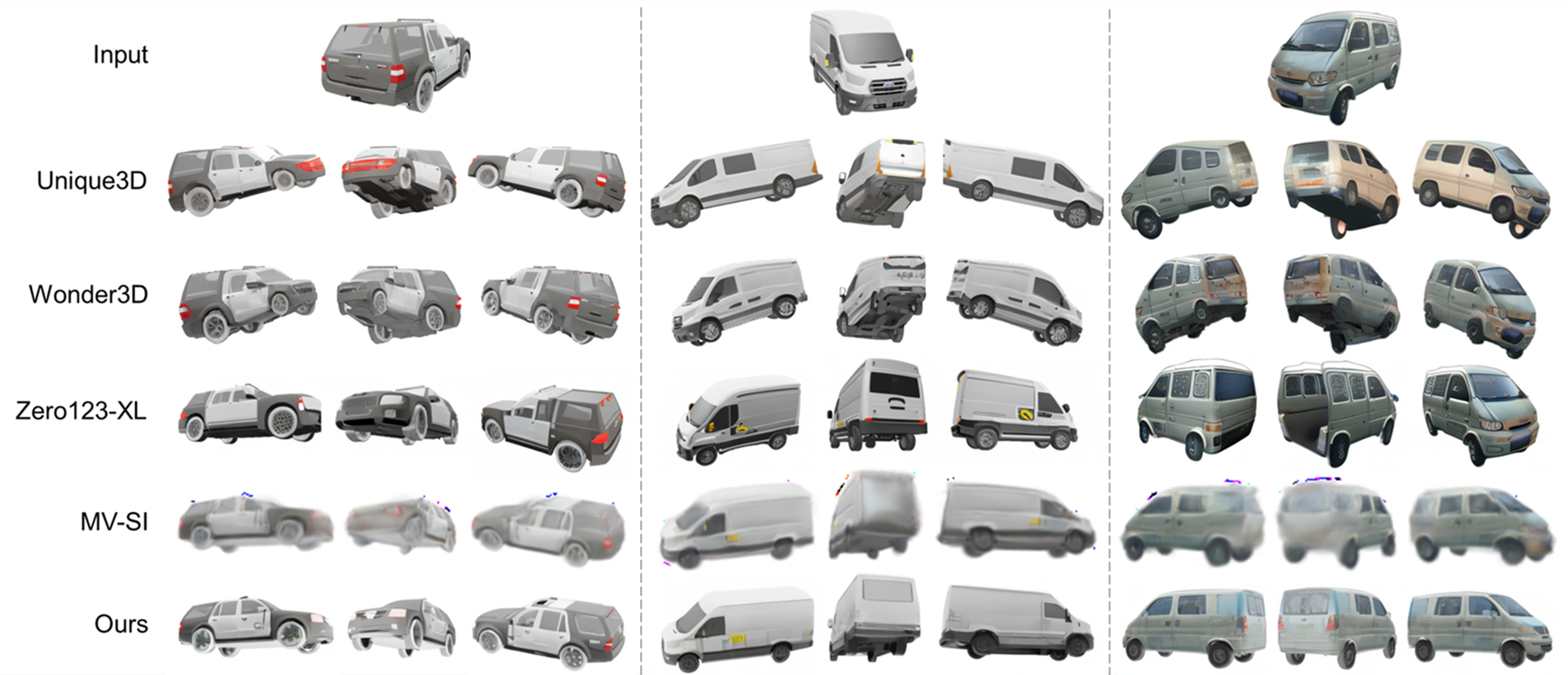}
		\end{center}
		\vspace{-4mm}
		\caption{Qualitative comparison of novel view synthesis results with baseline methods. The three input images are from SRN-Cars, SketchFab-Cars, and 3DRealCar datasets (left to right).
		}
		\label{fig:nvs_qualititave}
		\vspace{-4mm}
	\end{figure*}
	
	\noindent \textbf{Geometry refinement with joint optimization of color and normal.}
	ISOMER only relies on high-resolution normal maps for geometry refinement. 
	In fact, multi-view RGB images can also provide useful cues for shape optimization~\cite{worchel:2022:nds}.
	However, since RGB textures are much more complex and non-smooth than normal maps, when the texture is inconsistent across multiple views, it will significantly reduce the accuracy of reconstruction.
	In our work, both the generated color images and normal maps exhibit strong geometric consistency across viewpoints, allowing us to incorporate color constraints together with normal supervision to further refine the mesh geometry.
	To mitigate potential cross-view inconsistencies, we adopt the Explicit Target (ET) strategy~\cite{wu2024unique3d}, which aggregates observations on the mesh surface and re-renders them to each view for stable multi-view supervision.
	The final loss function is defined as:
	\begin{equation}
		\mathcal{L}_{refine} =  \mathcal{L}_{coarse} +  \mathcal{L}_{color}^{ET} +  \mathcal{L}_{normal}^{ET}.
		\label{eq:placeholder}
	\end{equation}


	\noindent \textbf{Texture projection.}
	To obtain high-quality textures, we perform vertex colorization using the synthesized dense color images.
	Here, we propose a one-shot color projection method to colorize the mesh vertices.
	First, all the multi-view images are assigned proper priorities based on their relative angles to the input view.
	Then the color images are processed one by one from highest to lowest priority, projecting pixel values onto the mesh vertices. Each vertex can be assigned a color only once. After traversing all views, the colorization process is complete.
	Benefiting from the dense and geometrically consistent texture images generated in the first stage, the mesh surface is well covered, producing clean textured meshes without additional color propagation.
	
	\section{Experiments}

	\subsection{Experimental Setting}

	\noindent \textbf{Datasets.} 
	The training data for MV-SI and the multi-view diffusion model consist of $2,811$ car models from ShapeNet-SRN (SRN-Cars)~\cite{sitzmann2019scene} and $1,045$ car models from Objaverse~\cite{deitke2023objaverse}. 
	For each object, we first normalize it to be centered and of unit scale. 
	To accommodate real-world images captured by perspective cameras with varying intrinsics, the input images are rendered with a randomly sampled  FOV  in the range of  $[5^\circ, 55^\circ]$. 
	For supervision, we render $30$ ground-truth (GT) images per object with camera elevation and azimuth randomly sampled from $[-20^\circ, 45^\circ]$ and $[0^\circ, 360^\circ]$, respectively. 
	Following Zero123++~\cite{shi2023zero123++}, the multi-view GT outputs are rendered with a fixed FOV of $30^\circ$.
	For evaluation, we use $50$ car models randomly selected from the test set of SRN-Cars and $50$ car models collected from Sketchfab\footnote{https://sketchfab.com} (Sketchfab-Cars).
	Moreover, we also evaluate visual quality on in-the-wild images collected from the 3DRealCar~\cite{du20243drealcar} dataset to show the generalization ability and practicality of our method.

	
	
	\noindent \textbf{Implementation details.}   
	We train MV-SI and the multi-view diffusion model separately.
	For MV-SI, we train from scratch using images with a resolution of $128\times 128$. It takes $6$ multi-view images as input and renders $8$ target images per iteration.
	The input images for MV-SI are generated using the Zero123++ model fine-tuned in~\cite{xu2024instantmesh}.
	After training, the MV-SI weights are fixed and used as a 3D-consistent conditioning for our multi-view diffusion model.
	It is initialized with the pretrained weights of Wonder3D~\cite{long2024wonder3d}.
	During fine-tuning, the model generates $6$ multi-view color images and normal maps simultaneously with specified viewpoints. 
	During inference, we generate three sets of multi-view images with elevations of $[0^\circ, -30^\circ, 30^\circ]$, and azimuths of $[0^\circ, 45^\circ, 90^\circ, 180^\circ, -90^\circ, -45^\circ]$. In every generation cycle, the pose of the first view is consistent with the first output of Zero123++. Together, we have $16$ images for the mesh reconstruction.


	\noindent \textbf{Baselines.} 
	We conduct comprehensive comparisons with three categories of baseline methods: 
	1) novel-view synthesis models: Zero123~\cite{liu2023zero},  Zero123-XL~\cite{liu2023zero} and Zero123++~\cite{shi2023zero123++}.
	2) general single-view generation: One-2-3-45~\cite{liu2023one}, Wonder3D~\cite{long2024wonder3d},  Unique3D~\cite{wu2024unique3d}, InstantMesh~\cite{xu2024instantmesh} and 3D-Adapter~\cite{chen20243d}). 3) car-specific generation: DreamCar~\cite{du2024dreamcar} and Splatter Image~\cite{szymanowicz2024splatter}.
	
	
	
	\noindent \textbf{Metrics.} 
	We investigate the performance of two tasks in 3D car generation: novel view synthesis (NVS) and image-to-3D mesh generation.
	For NVS evaluation, we compare the generated multi-view images with corresponding GT images and adopt PSNR, SSIM\cite{wang2004image}, and LPIPS~\cite{zhang2018unreasonable} as the metrics. 
	For the evaluation of 3D mesh, we first perform coordinate alignment between the generated meshes and their ground truth counterparts and then normalize them into a unit cube. The 3D reconstruction quality is evaluated by the Chamfer Distance (CD) and Volume Intersection of Union (VIoU).

	
	
	
	\subsection{Evaluation of Experiment Results}
	
	\begin{figure*}[tbp]
		\begin{center}
			\includegraphics[width=0.9\linewidth]{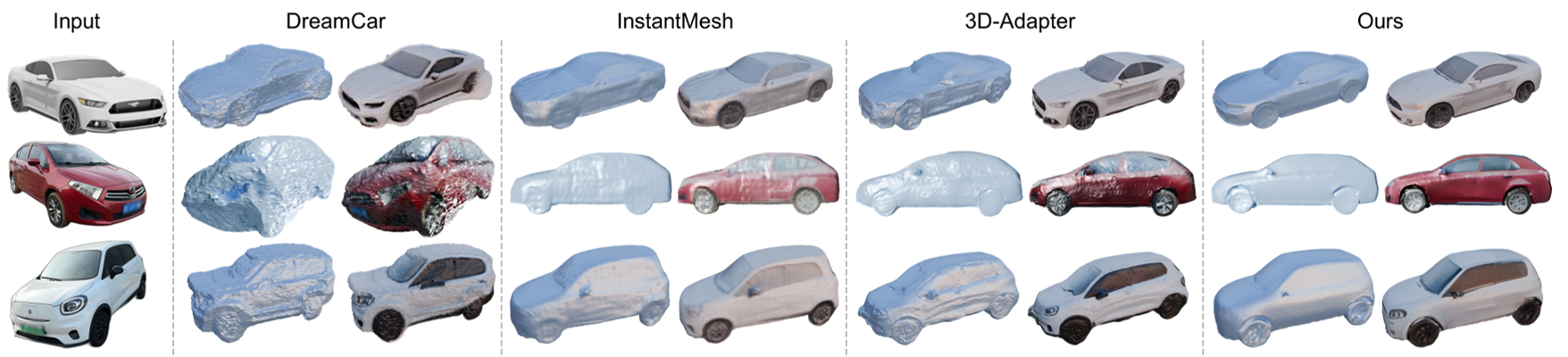}
		\end{center}
		\vspace{-6mm}
		\caption{Qualitative comparisons with baseline methods in 3D reconstruction.
		}
		\label{fig:mesh_qualititave}
		\vspace{-4mm}
	\end{figure*}

	\noindent \textbf{Evaluation of Novel view synthesis.}
	The quantitative comparison  is presented in Table~\ref{tab:nvs}. 
	Our method achieves the best overall performance on both datasets, especially in terms of LPIPS, demonstrating improved perceptual quality and cross-view consistency. 
	We also include the coarse renderings produced by MV-SI for comparison. 
	MV-SI achieves higher PSNR and SSIM scores due to its well camera-pose alignment with the GT images, but its smooth and slightly blurred renderings may further inflate these pixel-based metrics~\cite{zhang2018unreasonable}.
	Qualitative comparisons  are shown in Fig.~\ref{fig:nvs_qualititave}. 
	Unique3D~\cite{wu2024unique3d} and Wonder3D~\cite{long2024wonder3d} often produce distorted views when the input images are non-frontal and perspective. 
	Zero123-XL~\cite{liu2023zero} generates visually plausible images but lacks multi-view consistency since each view is synthesized independently, making it difficult to maintain accurate alignment with camera poses.
	In contrast, both MV-SI and our method exhibit strong geometric consistency across views. 
	After refinement by the multi-view diffusion model, 
	the coarse MV-SI renderings are transformed into clear and texture-rich images. 
	Moreover, our method can cope with car images with diverse  camera intrinsics and extrinsics, and generate 3D-consistent multi-view images from arbitrary viewpoints.

	\begin{table}[tbp]
		\centering
		\caption{Quantitative comparison in 3D reconstruction. 
			For fair comparison, we apply our ISOMER+ algorithm on the multi-view outputs from several baselines, indicated with $^\ast$. 
		}
		\vspace{-3mm}
		\label{tab:mesh}
		\setlength{\tabcolsep}{4pt}
		\begin{tabular}{cccccc}
			\toprule
			\multirow{2}{*}{Method} & \multicolumn{2}{c}{SRN-Cars}  & \multicolumn{2}{c}{SketchFab-Cars} & \multirow{2}{*}{Runtime} \\
			\cmidrule(lr){2-3} \cmidrule(lr){4-5} 
			& CD$\downarrow$ & VIoU$\uparrow$  & CD$\downarrow$ & VIoU$\uparrow$ & \\ 
			\midrule
			DreamCar~\cite{du2024dreamcar}  &  0.0401   &  0.3649  &   0.0399  &    0.3445 & 218.7min \\
			One-2-3-45~\cite{liu2023one} &   0.0409 &  0.4002 &   0.0626 &  0.2885 & 54.6s \\
			Unique3D~\cite{wu2024unique3d}  &     0.0323  &   0.3230 &    0.0285  &    0.3454 & 31.5s \\
			Wonder3D~\cite{long2024wonder3d}   & 0.0255    & 0.4388  &   0.0323  &    0.3582 & 249.6s \\
			Wonder3D$^\ast$~\cite{long2024wonder3d}   &   0.0281  &   0.4182  &  0.0298 &  0.3637  & 29.3s \\
			Zero123++$^\ast$~\cite{shi2023zero123++}  &   0.0166  & 0.4012 &   0.0153  &   0.4141  & 27.1s\\
			InstantMesh~\cite{xu2024instantmesh} &  0.0154 &  0.4367  &   0.0177   &  0.4321 & 9.6s \\
			3D-Adapter~\cite{chen20243d}  &   \cellcolor{yellow!30} 0.0128 & \cellcolor{red!15}  0.4738   &  \cellcolor{yellow!30}  0.0117   &   \cellcolor{yellow!30} 0.4440 &  267.5s \\
			\midrule
			Ours   & \cellcolor{red!15} 0.0114  & \cellcolor{yellow!30} 0.4643   &  \cellcolor{red!15} 0.0098  &  \cellcolor{red!15} 0.4548 & 138.4s\\ 
			\bottomrule
		\end{tabular}
		\vspace{-5mm}
	\end{table}

	\noindent \textbf{Evaluation of Single image-to-3D generation.}
	Table~\ref{tab:mesh} presents  the quantitative comparison of 3D reconstruction quality. 
	Our method achieves superior performance on both datasets and significantly improves geometry accuracy compared with our baseline methods, i.e., Zero123++~\cite{shi2023zero123++} and Wonder3D~\cite{long2024wonder3d}. 
	For fair comparison, we apply the same mesh reconstruction algorithm (ISOMER+) to the multi-view outputs of these methods, indicating that their synthesized views are inferior to reconstruct accurate meshes.  
	Meanwhile, our method maintains competitive runtime while achieving superior reconstruction quality.
	Qualitative comparisons are shown in Fig.~\ref{fig:mesh_qualititave}. 
	DreamCar~\cite{du2024dreamcar} struggles with single-image inputs and often fails to recover fine geometric details. 
	InstantMesh~\cite{xu2024instantmesh} performs better in geometry, yet it can only synthesize blurry and low-resolution textures. 
	3D-Adapter~\cite{chen20243d} also adopts synchronization during the generation process, but the reconstructed meshes suffer from unstable geometry, with noticeable surface artifacts and distorted body shapes.
	In contrast, our method produces more accurate geometry and clearer textures, resulting in smoother and more consistent 3D car models.

	\subsection{Ablation Study}

	We conduct ablation studies on the SketchFab-Cars dataset 
	to better understand the contributions of each component.
	
	\noindent \textbf{Ablation for multi-view diffusion model.}
	Table~\ref{tab:ablation_mv_diffusion} presents an ablation study on the training strategy and key components of our multi-view diffusion framework.
	Starting from the Wonder3D baseline (ID1), fixed-view fine-tuning (FV-f.t., ID2) improves performance by adapting the model to the target data distribution.
	In contrast, performing arbitrary-view fine-tuning (AV-f.t., ID3)  individually  leads to degraded results due to the lack of geometric synchronization between views. 
	Adding MV-SI (ID$4$) significantly improves the performance, indicating the effectiveness of geometric synchronization for multi-view generation.
	Finally, further incorporating the multi-view SDEdit sampling strategy (ID$5$, ours) achieves the best overall performance, demonstrating improved cross-view consistency and visual fidelity.

	\begin{table}[tbp]
		\centering
		\caption{Ablation on the multi-view diffusion design. 
		}
		\vspace{-3mm}
		\renewcommand{\arraystretch}{0.85}
		\label{tab:ablation_mv_diffusion}
		\setlength{\tabcolsep}{1pt}
		\begin{tabular}{c|cccc|ccc}
			\toprule
			ID & FV-f.t. & AV-f.t. & MV-SI & mvSDEdit & PSNR$\uparrow$ & SSIM$\uparrow$ & LPIPS$\downarrow$   \\
			\midrule
			
			1 (Wonder3D) &  & & & & 19.48    &  0.835 &   0.116 \\
			2  & \textcolor{black}{\ding{51}} & & & & 20.18    &  0.842 &   0.107 \\
			3  &  & \textcolor{black}{\ding{51}} & & & 18.88    &  0.804 &   0.130\\
			4  & & \textcolor{black}{\ding{51}} & \textcolor{black}{\ding{51}} &  & 	20.80	 &	0.842	 &	  \textbf{0.092 } \\
			5 (Ours) & & \textcolor{black}{\ding{51}} & \textcolor{black}{\ding{51}} &  \textcolor{black}{\ding{51}}  & \textbf{21.58} & \textbf{0.855} & 0.096    \\
			\bottomrule
		\end{tabular}
	\end{table}

	\begin{table}[!t]
		\centering
		\caption{ Ablation on ISOMER+.
		}
		\vspace{-3mm}
		\renewcommand{\arraystretch}{0.9}
		\label{tab:ablation_isomer_plus}
		\setlength{\tabcolsep}{2pt}
		\begin{tabular}{c|ccc|c}
			\toprule
			ID & \makecell{Sphere \\ Initialization} & $\mathcal{L}_{Laplace}$ &  $\mathcal{L}_{color}^{ET}$ & CD$\downarrow$   \\
			\midrule
			1 (ISOMER) &  & &  &   0.08682 \\
			2 & \textcolor{black}{\ding{51}} &  & 	&  0.01051 \\
			3 &  \textcolor{black}{\ding{51}} &  \textcolor{black}{\ding{51}}  &   & 0.01044  \\
			4 (ISOMER+) &  \textcolor{black}{\ding{51}} &  \textcolor{black}{\ding{51}}  &   \textcolor{black}{\ding{51}} &  \textbf{0.01029}  \\
			\bottomrule
		\end{tabular}
		\vspace{-5mm}
	\end{table}

	\begin{figure}[!t]
		\begin{center}
			\includegraphics[width=0.8\linewidth]{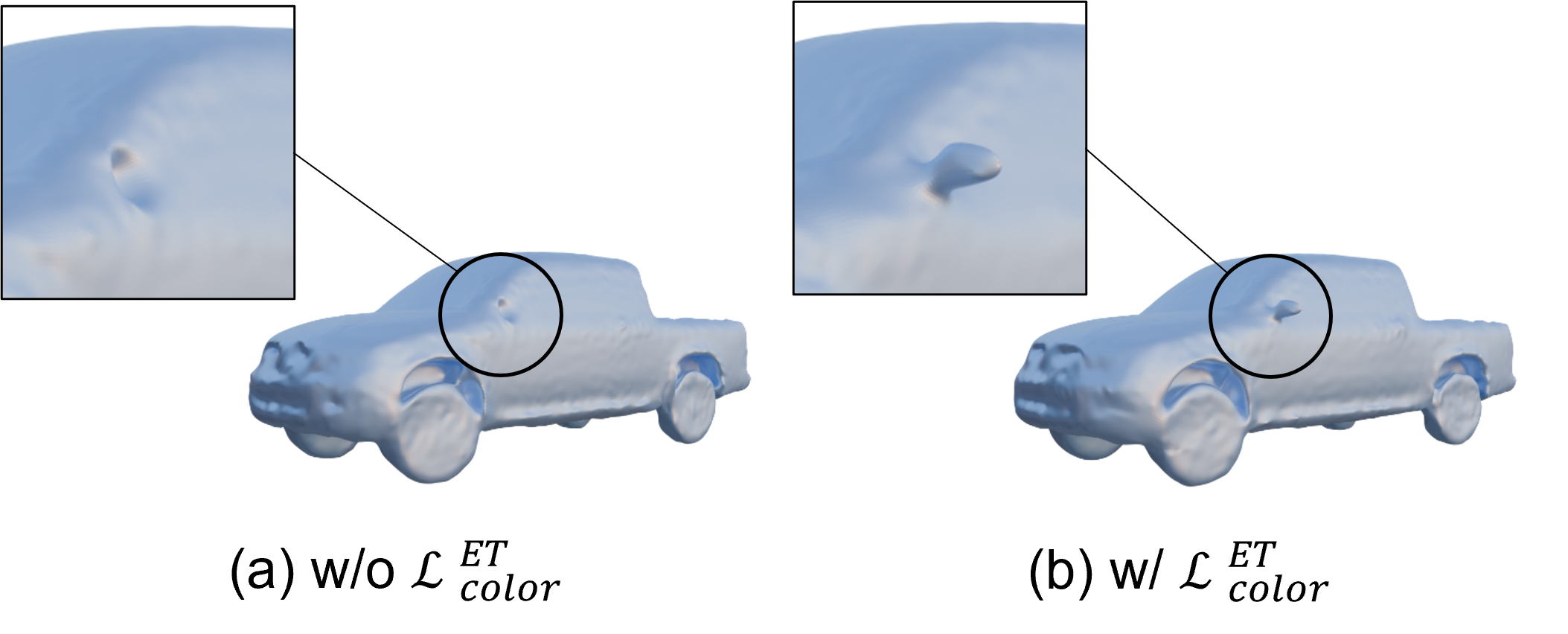}
		\end{center}
		\vspace{-5mm}
		\caption{Ablation study on $\mathcal{L}_{color}^{ET}$ in ISOMER+. 
		}
		\label{fig:wo_w_RGB_loss_comparison}
		\vspace{-5mm}
	\end{figure}

	\noindent \textbf{Ablation for 3D reconstruction.}
	A detailed ablation on ISOMER+ is shown in Table~\ref{tab:ablation_isomer_plus}.
	The results show that each component plays a complementary role in improving the overall reconstruction quality.
	Fig.~\ref{fig:wo_w_RGB_loss_comparison} compares the effectiveness of color optimization ($\mathcal{L}_{color}^{ET}$) in the refinement stage.
	Specifically, without $\mathcal{L}_{color}^{ET}$ loss (left), the car models struggle to accurately reconstruct fine geometric features, such as side mirrors, resulting in artifacts or missing details. In contrast, adopting $\mathcal{L}_{color}^{ET}$ loss (right) can help recover the accurate fine details of car models. 

	\section{CONCLUSIONS}
	
	In this work, we present 3DCarGen, a  framework for generating high-quality 3D car models from a single real-world image. 
	By incorporating a geometry-consistent multi-view diffusion model and a robust mesh reconstruction stage, our approach effectively addresses the limitations of existing methods, such as view inconsistency, limited viewpoints, and suboptimal mesh quality. 
	Compared with baseline methods, 3DCarGen achieves superior performance in generating 3D-consistent views and high-fidelity 3D car assets.  
	This work presents a practical step towards scalable 3D car generation for autonomous driving and digital content creation.
	


	


	\section*{APPENDIX}
	\subsection{More Implementation Details}      
	\noindent \textbf{Training details of MV-SI.} 
	MV‑SI's backbone follows SongUNet~\cite{song2020denoising} except for two key modifications: (i) The final layer is replaced with a $1\times1$ Conv to predict per‑pixel 3D Gaussians. (ii) For multiple inputs, positional embeddings and cross-attention layers are applied to allow information interaction between views. 
	The training process consists of two stages: the first stage runs for $80$K steps with a learning rate of $5\times10^{-5}$, and the second stage continues for an additional $60$K steps with a reduced learning rate of $1\times10^{-5}$. Both stages use a batch size of $24$. 
	During each training iteration, MV-SI receives 6 conditioning images to construct a low-resolution Gaussian representation and render 8 target views. 
	The view inconsistencies introduced by Zero123++ are effectively eliminated through cross-attention.
	In total, the training of MV-SI takes approximately 7 days using 6 A100 40GB GPUs.
	
	\noindent \textbf{Training details of multi-view diffusion model.} 
	The multi-view diffusion model is initialized with the weights from Wonder3D~\cite{long2024wonder3d}.
	We train it using $256\times256$ resolution images, with a total batch size of $256$, for $20$K steps. Due to computational constraints, the model is trained to generate 6 target views for each generation. The training process takes approximately 3 days on 32 A100 40GB GPUs.

	\noindent \textbf{Clarification on camera poses.}
	In our settings, the pose of the input image is unknown. Relative poses refers to the azimuth transformation of the target views relative to the input image.
	The azimuth angles of the multi-view images generated by Zero123++~\cite{shi2023zero123++} are relative to the input image, and the elevation angles are absolute. The 3D-consistent multi-view diffusion model also follows this setting.
	We pass the relative azimuths and absolute elevations of the target views to the MV diffusion model via camera embeddings to enable it to synthesize novel views at specified angles.

	\subsection{More Visualization Results}      
	
\begin{figure}[htbp]
		\begin{center}
			\includegraphics[width=0.9\linewidth]{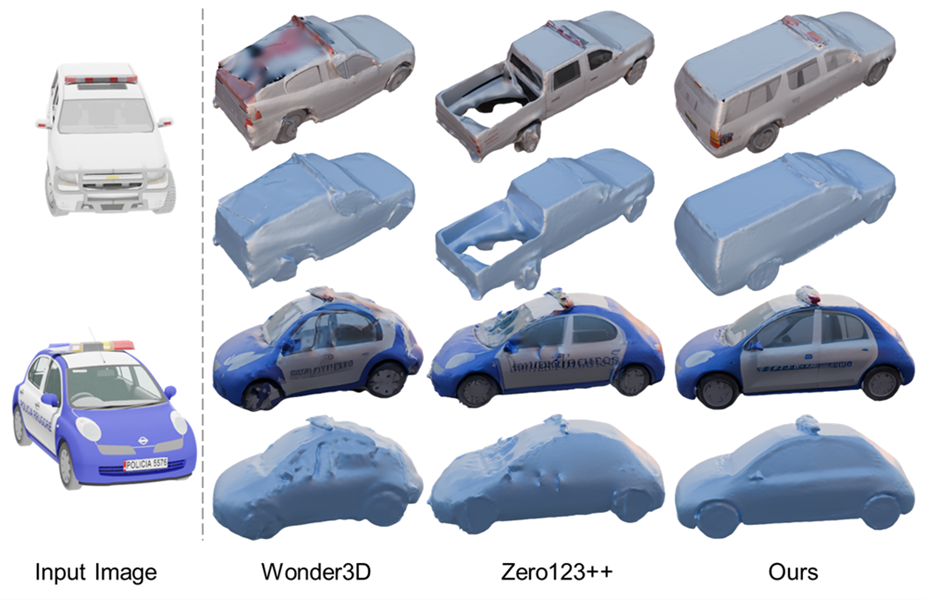}
		\end{center}
		\vspace{-5mm}
		\caption{Qualitative comparison of 3D reconstruction using ISOMER+ on multi-view images generated by different methods.}
		\label{fig:sup_inconsistency}
		\vspace{-3mm}
	\end{figure}

	Fig.~\ref{fig:sup_inconsistency} presents a qualitative comparison of multi-view 3D reconstruction using different methods.  Wonder3D~\cite{long2024wonder3d} suffers from geometric distortions, and Zero123++~\cite{shi2023zero123++} fails to maintain structural stability across views. In contrast, our method demonstrates superior multi-view consistency, ensuring stable geometry and texture across viewpoints. 
	
\begin{figure}[htbp]
		\begin{center}
			\includegraphics[width=0.9\linewidth]{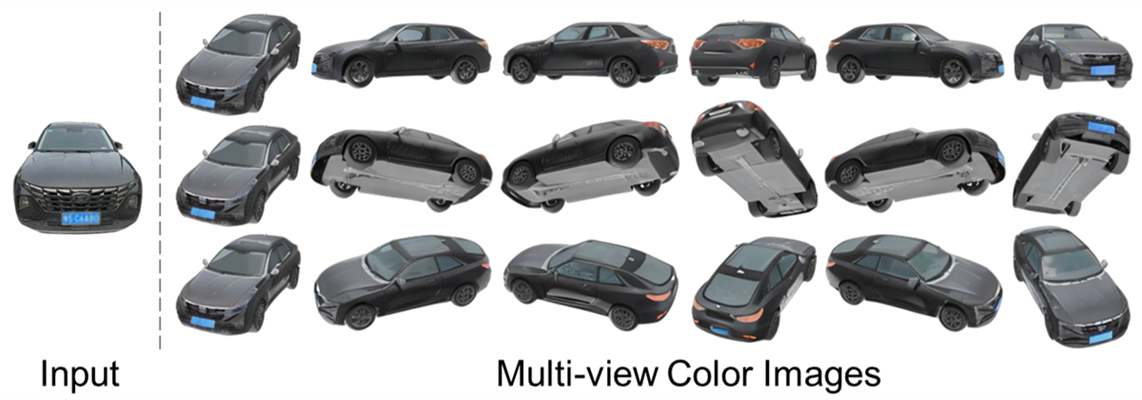}
		\end{center}
		\vspace{-5mm}
		\caption{
			Example of multi-view RGB images generated by our method from a single input image. 
		}
		\label{fig:supp_nvs_ours}
		\vspace{-3mm}
	\end{figure}

	Fig.~\ref{fig:supp_nvs_ours} presents more multi-view color images generated by 3DCarGen from a single car image. 
	The synthesized views cover diverse viewpoints while maintaining strong geometric consistency across views, preserving the car's shape, texture, and structural integrity across various viewpoints. 

\begin{figure}[htbp]
		\begin{center}
			\includegraphics[width=0.9\linewidth]{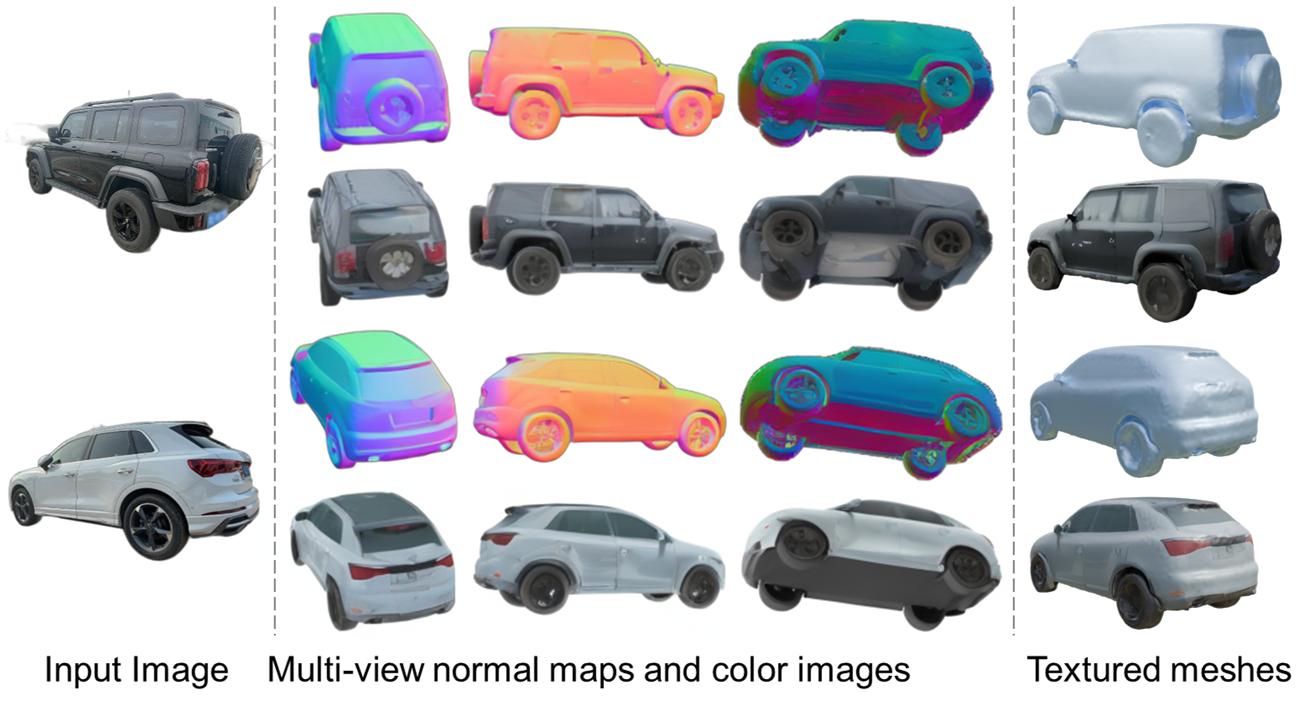}
		\end{center}
		\vspace{-6mm}
		\caption{
			More results of our method on real-world car images. }
		\label{fig:our_results}
		\vspace{-6mm}
	\end{figure}

	Fig.~\ref{fig:our_results} shows additional qualitative results of our method on real-world car images. 
	Given a single input image, our framework generates geometrically consistent  multi-view color images and normal maps from diverse viewpoints and reconstructs high-quality textured meshes. 
	
	\bibliographystyle{IEEEtran}
	\bibliography{IEEEtranBST/ref}

\end{document}